# Distributed Deep Learning for Precipitation Nowcasting

Siddharth Samsi, Christopher J. Mattioli, Mark S. Veillette
MIT Lincoln Laboratory, Lexington, MA

***Abstract*—Effective training of Deep Neural Networks requires massive amounts of data and compute. As a result, longer times are needed to train complex models requiring large datasets, which can severely limit research on model development and the exploitation of all available data. In this paper, this problem is investigated in the context of precipitation *nowcasting*, a term used to describe highly detailed short-term forecasts of precipitation and other hazardous weather. Convolutional Neural Networks (CNNs) are a powerful class of models that are well-suited for this task; however, the high resolution input weather imagery combined with model complexity required to process this data makes training CNNs to solve this task time consuming. To address this issue, a data-parallel model is implemented where a CNN is replicated across multiple compute nodes and the training batches are distributed across multiple nodes. By leveraging multiple GPUs, we show that the training time for a given nowcasting model architecture can be reduced from 59 hours to just over 1 hour. This will allow for faster iterations for improving CNN architectures and will facilitate future advancement in the area of nowcasting.**

## I. Introduction

Deep Neural Networks (DNN) have been successfully applied in many diverse domains such as image classification, video analysis, language modeling and translation, medical imaging and weather [1]–[5]. Recently, there has been increasing interest using DNNs to generate and improve weather forecasting [6], [7], [8], [9]. Of particular interest is in the area of *nowcasting*, a term used to describe high resolution, short-term (e.g. 0 to 2 hours), weather forecasts of precipitation or other meteorological quantities. In contrast to larger numerical weather prediction (NWP) models (which may take hours to run), nowcasts typically take on the order of minutes to generate a forecast. Because of this, they can update frequently using recent observations. The low data latency and fast processing time make nowcasts generally more accurate than NWP models for short lead times [10]. These types of forecasts are widely used in public safety, air traffic control, tactical mission planning and many other areas where high fidelity and rapidly updating forecasts are needed.

There are a number of operational nowcasting systems running, including the Corridor Integrated Weather System (CIWS, [11]), the Auto-nowcaster [12] and the Multi-radar multi sensor [13] (MRMS) system. Each of these systems use a suite of image processing algorithms to generate their forecasts, which have required development over several years. For this reason, the use of DNNs for nowcasting is appealing because they could enable a self-learning, data-driven methodology for creation of these forecasts, and would remove much of the hand-tuning required to develop these algorithms. The challenges with implementing a DNN for this problem include dealing with the high image resolutions of the input data, as well as the complexity required to effectively model evolving weather.

This material is based upon work supported by the Assistant Secretary of Defense for Research and Engineering under Air Force Contract No. (FA8721-05-C-0002 and/or FA8702-15-D-0001). Any opinions, findings and conclusions or recommendations expressed in this material are those of the author(s) and do not necessarily reflect the views of the Assistant Secretary of Defense for Research and Engineering.

The effective training of DNNs requires the use of large amounts of labelled data. As computational power has increased, models have gotten deeper and wider, leading to a corresponding increase in the time required to train them. For example, the VGG16 [14] model has almost 140 million trainable parameters. Other published networks have similarly large numbers of parameters and can take days to train. A common approach to reducing the training time for DNNs is the use of data parallelism [15]–[17]. In this approach, the model is duplicated on multiple machines, with each machine training on a subset of the data. Gradients are calculated and shared across all machines and weights are updated accordingly. An alternative to this is the model parallel approach where the model itself is distributed across multiple machines [15], [18]. This method is typically beneficial in situations where the model is too large to be held in the machine's memory. Additionally, models with large numbers of parameters or computations per layer can benefit from this approach [15].

In this paper, we present a performance study on applying the data parallel approach to training a nowcasting model. We use a convolutional neural network as described in Section II to predict future images of precipitation given a sequence of past images. The model was developed in TensorFlow/Keras [19] and the data parallelism was achieved using the Horovod [20] framework. Benchmarking was performed on two datasets to study the effect of training data size on the scalability of the model training process.

## II. Deep Learning for Nowcasting

This section will describe the training data and model architecture used for the nowcasting model. While the resulting performance of the nowcasting model still falls short of state-of-the-art, it is a useful architecture for testing and demonstrating scaling properties of distributed learning (which is the primary focus of this work). Lessons learned and future steps on how to improve the model will be discussed.

### A. Previous Work

There has already been a number of attempts of using deep learning for precipitation nowcasting. Convolutional Long Short Term Memory (ConvLSTM) layers were developed for this purpose in [6], which demonstrated that DNNs are capable of generating forecasts. In [21], additional model architectures were explored, and distributed learning was also applied to train models and speed up hyperparameter searches (although benchmarking performance was not a focus). The DNN model trained in this work will differ from those considered previously in that they will not contain recurrent layers, and they will be capable of generating forecasts on arbitrarily sized grids by using a fully convolutional architecture.

### B. Data Description

The input to the nowcasting CNN is a sequence of single channel images that depict precipitation intensity. In this work, precipitation is represented using Vertically Integrated Liquid (VIL) [22], which measures the amount of liquid water content in the atmosphere above a particular point, and is estimated from weather radar measurements. In the CIWS system, the Level III VIL product generated around each WSR-88D (NEXRAD) radar using the Open Radar Product Generator (ORPG, [23]) are combined into a single national mosaic. The pixels in this image represent a 1 km by 1 km area, and the CIWS grid is 3520 by 5120 pixels in size. VIL is normalized to the range [0,255] using the "digital VIL" transformation [5]. These normalized images will provide the training data for the CNN trained in this work.

Because sequences of national sized CIWS images would be too large to be processed in GPU memory, the CNN in this work is trained using smaller "patches" of VIL sampled randomly from the CIWS grid. The process for generating a training dataset is summarized as follows: First, a set of random times between Jan 1, 2017 and Dec 30, 2018 are chosen. For each random time $t_0$, a sequence $S_{t_0} = \{I_{t_0-60}, I_{t_0-50}, \ldots, I_{t_0}, \ldots, I_{t_0+60}\}$ of CIWS images is gathered that is temporally centered at $t_0$ and separated by 10 minutes. A set of spatial coordinates $(x_i, y_i)$ are randomly selected from within $I_{t_0}$ to be used as patch centers. These points are selected only in regions within weather radar range (230 km around each radar) and areas in $I_{t_0}$ with heavier precipitation were sampled with higher likelihood to avoid over sampling cases with no precipitation. Around each point $(x_i, y_i)$, a 256x256 patch is extracted from each image in the sequence $S_{t_0}$. The "past" patches from $I_{t_0-60}, \ldots, I_{t_0}$ are used as inputs to the model, and the "future" patches from $I_{t_0+10}, \ldots, I_{t_0+60}$ are used as truth during training. Repeating this process over all selected times results in a tensor $X$ of size $[N, 256, 256, 7]$, where $N$ is the total number of collected patches to be used as input to the CNN, with "truth" $Y$ of size $[N, 256, 256, 6]$. See Figure 1. All patches are then normalized to have zero mean and unit variance.

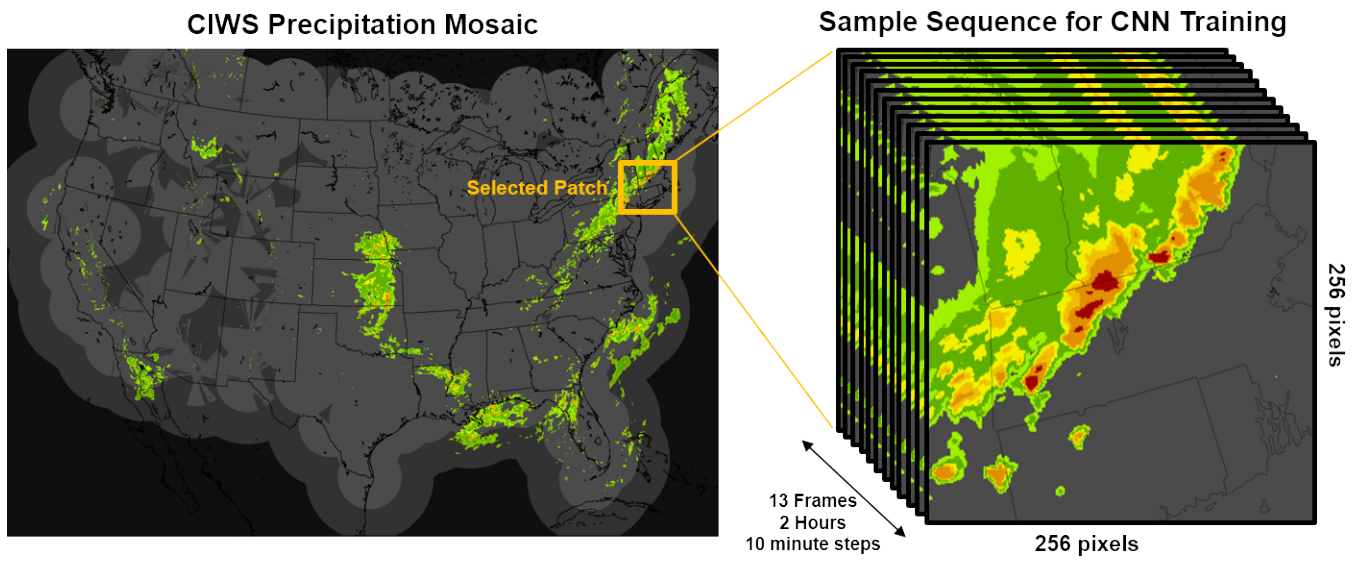

Fig. 1: To collect training data for the nowcast, patches of size 256km x 256km (right) are selected randomly from the CIWS national mosaic (left). For each selected patch, image data spanning two hours is sampled to create a training sample. The images with times before and including the target time are used as inputs to a CNN, and the images after the target times are used as truth during training.

### C. Model Architecture

The model architecture is shown in Figure 2. It contains an encoder/decoder similar to the U-net architectures [24], along with forecasting layers attached to the outputs of the decoding steps. In the encoding phase, the input of size 256x256x7 representing 7 frames at a 1 km resolution are processed by 4 convolutional layers with strides of 2. Each strided convolution effectively generates feature maps of lower resolutions, ending at 16km after the 4th convolution. In the decoding phase, each layer is upsampled to twice the size, and followed by another convolution. Skip layers are also used to connect encode and decode layers of similar spatial resolutions. This process of upsampling followed by convolution was chosen over deconvolution to avoid "checkerboarding" artifacts commonly observed with deconvolutional layers. At each stage of the decoding, an output at the current resolution is generated (this is represented in Figure 2 along the bottom). These lower resolution outputs are upsampled, combined with the output of the next decoding and then used to generate the output at the next resolution. This was done to allow the network to "build" forecasts from low resolution to high resolution. In the final output, three additional convolutions are applied to generate the final output at a 1 km resolution.

The CNN architecture used here is *fully convolutional* (no dense layers) and the convolutional layers do not utilize any padding. This is done so that convolutional layers trained on patches will generalize to arbitrarily sized grids. This is essential for creating models that can be trained efficiently on small patches, and then applied on large grids in operations.

With the chosen filter sizes, the "final" 1 km output of the CNN is an image of size 54x54x6. A loss function is applied to each of the outputs and summed with equal weights. These losses are only applied to the middle 48km x 48km section of each output. This is necessary to avoid edge artifacts caused by weather moving into and out of the scene. The loss function chosen for the architecture is mean squared error (MSE) between the predicted VIL estimates and truth downsampled to the necessary resolution. Because MSE is the only loss function used, it should be expected that the network generates images that are blurred with respect to the observed weather.

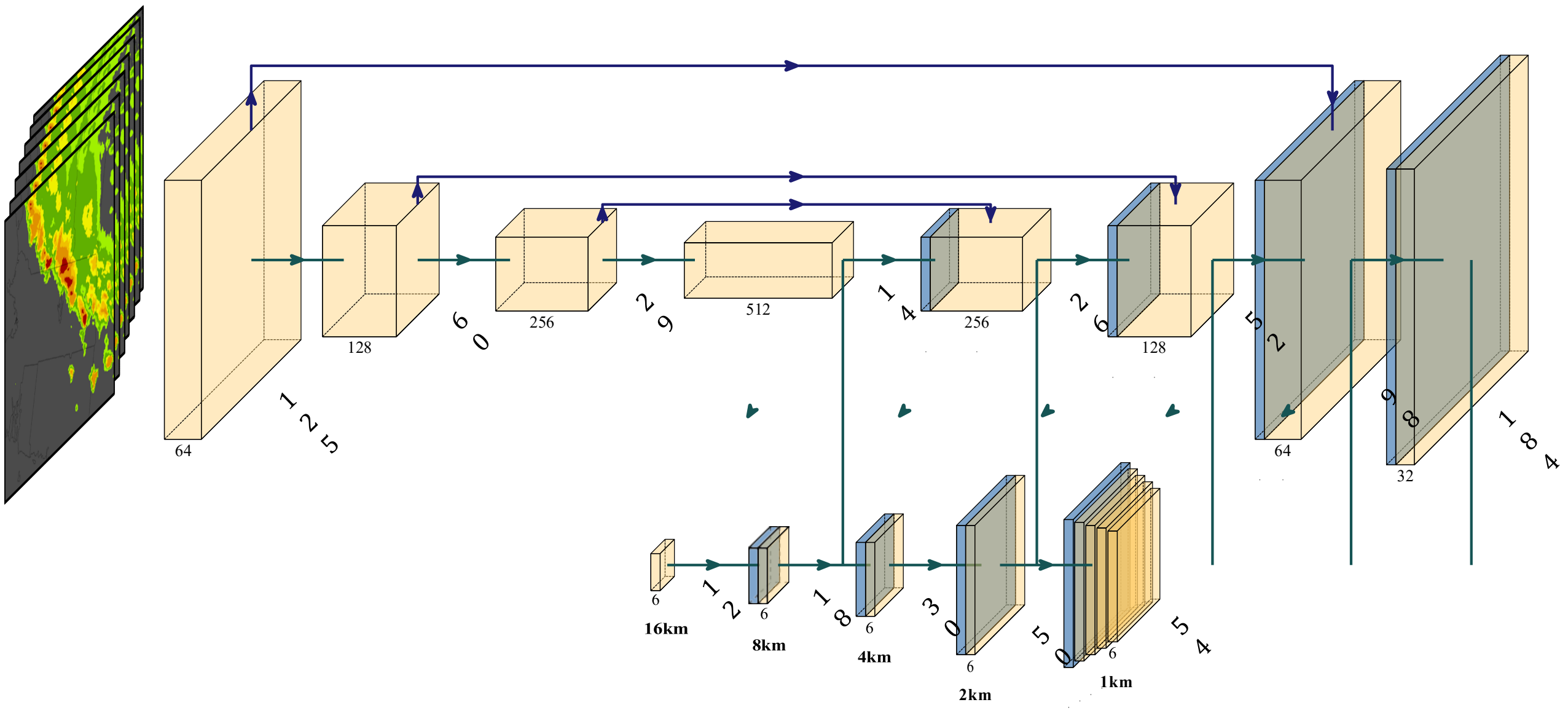

Fig. 2: Model architecture used for testing. This model takes as input a temporal sequence of 7 images and outputs the future 6 images. The model contains a encoding phase consisting of a sequence of convolutions (yellow tiles) that reduce the input spatial resolution to 1/16th the size. The decoding phase up-samples each layer (grey tiles) and applies additional convolutions. At each stage of the decoding, a forecast at that lower resolution is generated and compared to a down sampled version of truth. Skip layers are used to connect encoding layers to associated decoding layers.

## III. Experimental Design

Benchmarking experiments were performed on the Lincoln Laboratory Supercomputing Center (LLSC) TX-Green supercomputer. This is a heterogeneous system comprising a variety of hardware platforms from AMD, Intel and NVIDIA. The tests in this paper were performed on compute nodes with NVIDIA K80 GPUs. GPU nodes in the cluster consist of a dual socket Haswell (Intel Xeon E5-2680 v4 @ 2.40GHz) processor and two NVIDIA K80 GPUs. Each K80 GPU consists of two GK210 devices with 11.44 GB of GDDR5 memory each. Thus, a process running on these compute nodes sees four GPU devices.

### *A. Single GPU Training*

The nowcast model for weather forecasting was implemented in TensorFlow 1.12 using the Keras API. This network has 17,395,992 trainable parameters. While this number is orders of magnitude smaller than that of deeper networks such as VGG16 with almost 140 million parameters, the data sizes involved in training the nowcast model can make training time prohibitive. Initial studies were performed to understand the training performance on a single GK210 device. Two datasets were generated for training as described in Section II-B. One dataset consisted of 17,833 training images of size 256x256 pixels and 7 channels (Dataset I). A second dataset consisting of 45,897 images of the same size was used and included data from Dataset I. We will refer to this as Dataset II. In both cases, a testing set of 10,052 images distinct from all the training images was used. The model was trained on both datasets for 100 epochs using a single GK210 GPU with a batch size of 128. Larger batch sizes resulted in out-of-memory errors on the GPU. Table I summarizes the training time for the two datasets.

| | **Number of Images** | **Number of Epochs** | **Training Time (Hours)** |
|---|---|---|---|
| Dataset I | 17,833 | 100 | 23.219 |
| Dataset II | 45,897 | 100 | 59.136 |

Table I: Training time for the nowcast model on a single GK210 GPU. The model was trained for 100 epochs on two datasets, each with 10,052 test images.

These training times illustrate the challenges in the ability to explore different model architectures and hyperparameters. This limits not only the variety of model architectures that can be explored, but also limits the ability to train models on larger datasets. In order to address these challenges and enable the rapid prototyping of such models, we implemented the nowcast model using a multi-node, data distributed approach to training.

### *B. Multi-Node Distributed Training*

The model described in Section II-C was implemented in TensorFlow/Keras and was parallelized using the Horovod [20] framework. The Horovod framework was configured to use

OpenMPI for parallel communication. This enabled the parallelization of the model to leverage multiple GPUs on multiple nodes. The Horovod approach was evaluated on up-to 32 nodes with 4 GK210 devices per node, for a total of 128 GPU devices.

Data distributed training of the Nowcast model followed the approach described in [17]. In this approach, each of $N$ GPU devices load $1/N$ of the training dataset stored as an HDF5 file on a shared file system. In training iteration $t$, mini-batches $B_i$, $i = 1, \ldots, N$ of size $n$ are sampled from the device's data and processed by the network. The gradients computed on each device are then averaged as $\frac{1}{nN}\sum_i \sum_{x \in B_i} \nabla P(x, \omega_t)$ and used to update the global weights $\omega_t$ across all devices. To maintain consistent validation loss as more GPUs are used, the learning rate $\eta$ should be adjusted based on the number of devices. Following the empirical recommendations by Goyal et al., a learning rate of $\eta N$ following a gradual warm up phase lasting 5 epochs was found to result in good convergence behavior of the model on large numbers of GPUs. Based on experiments with a single GPU, the learning rate $\eta$ was set to 0.0002. Validation loss was computed on each device using a random 30% of the test images which were stored in a separate HDF5 file.

## IV. Results

### A. Selection of batch size

The batch size used for training neural networks can affect the training time and model accuracy. The selection of an appropriate batch size is dictated not only by the desired model accuracy but also the hardware capabilities such as device memory. For the nowcast model, we experimented with batch sizes of 8, 16, 32, 64 and 128 on both datasets. The training time for these batch sizes when using 8 GK210 devices (four K80 GPUs on two compute nodes) is shown in Figure 3. The training time for 100 epochs decreases as the batch size is increased. However, using a batch size of 128 results in a 4.5% increase in the time required to train the model for 100 epochs on Dataset I as compared with a batch size of 64. A per-device batch size of 128 is also seen to achieve a minimum validation loss of 3.0036 and 2.9769 for Datasets I and II, respectively. This comparison was performed on 8 GPU devices because 8 and 16 GPU devices were found to provide good parallel scaling for both datasets, as described in the next section.

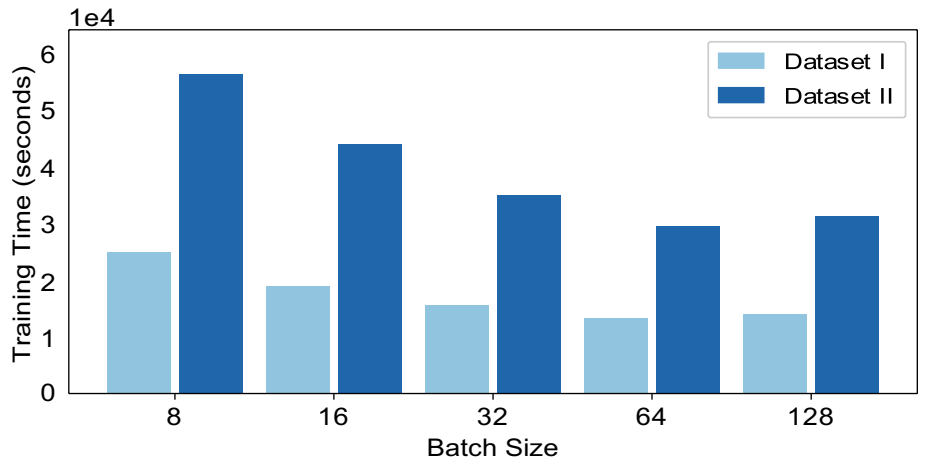

Fig. 3: Training time for the Nowcast model on 4 NVIDIA K80 GPUs for different batch sizes.

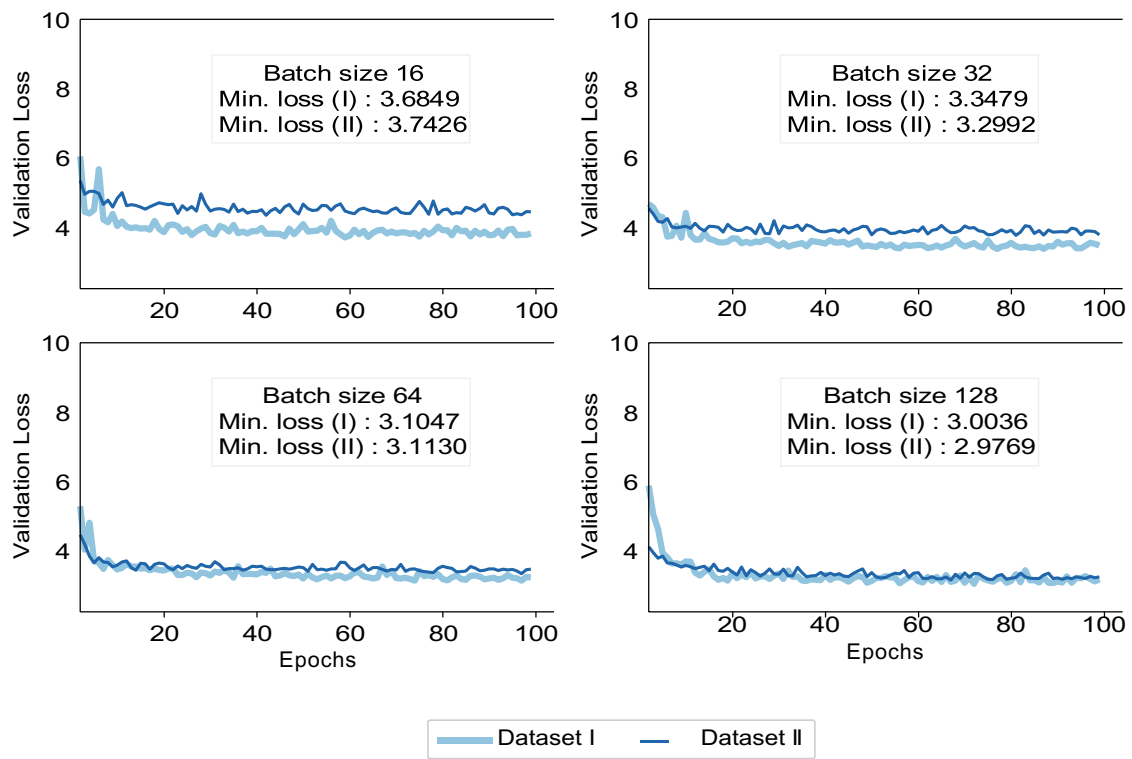

Fig. 4: Validation loss for the nowcast model on 4 NVIDIA K80 GPUs (corresponding to 8 GK210 GPU devices) for different batch sizes. First 3 epochs are not shown because the loss is higher in the initial epochs. In both datasets, the model achieves the smallest loss with a batch size of 128 per GK210 GPU device at the cost of a 4.5% increase in training time as shown in Figure 3.

### B. Distributed Training Performance

The wall time required for completing 100 epochs of training for the Nowcast model on multiple GPUs is shown in Figure 6. In terms of raw wall time, training the nowcast model on a single GK210 device takes 23.21 hours and 59.13 hours for Dataset I and II, respectively. By leveraging multiple GPUs, this can be reduced to just over an hour for both datasets. However, simply using 16 GPUs for either model reduces the total training time to 2.3 and 4.7 hours for Dataset I and II, respectively. This reduction in training time makes it feasible to train and test multiple models in a single day. Additionally, as we use larger datasets for training these types of models, we anticipate building better quality models without an exponential increase in the training time.

The nowcast model was trained on both datasets using up-to 128 GPUs. The training data was split across each GPU and a random subset of 30% of the test images were used by each GPU for testing and validation. The batch size used was 128 per device. Figure 5 shows the validation loss for each GPU count. As the number of devices used increased, the validation loss reduces smoothly, until $N = 24$. After this, the validation loss shows noisy behavior in the initial epochs. This can be attributed to the significant reduction in the number of training images available for each device. The significant variations in the validation loss can be attributed to the fact that the model tends to overfit locally on each GPU device and is not generalized well across the entire dataset. As a result of this, the gradient averaging step causes the model to adjust the weights significantly at each epoch. This behavior is more apparent when using 128 GPUs as seen in Figure 5. In this case, each GPU is only training on 139 images in case of Dataset I and 358 images for Dataset 2. Thus, scaling up to larger and larger number of devices is only beneficial up-to a certain $N$ and is also dependent on the amount of training data used. At the same time, when 4 to 48 GPUs are used to train the model, the larger effective batch size produces a lower minimum validation loss for both the datasets than is achieved with a single GPU.

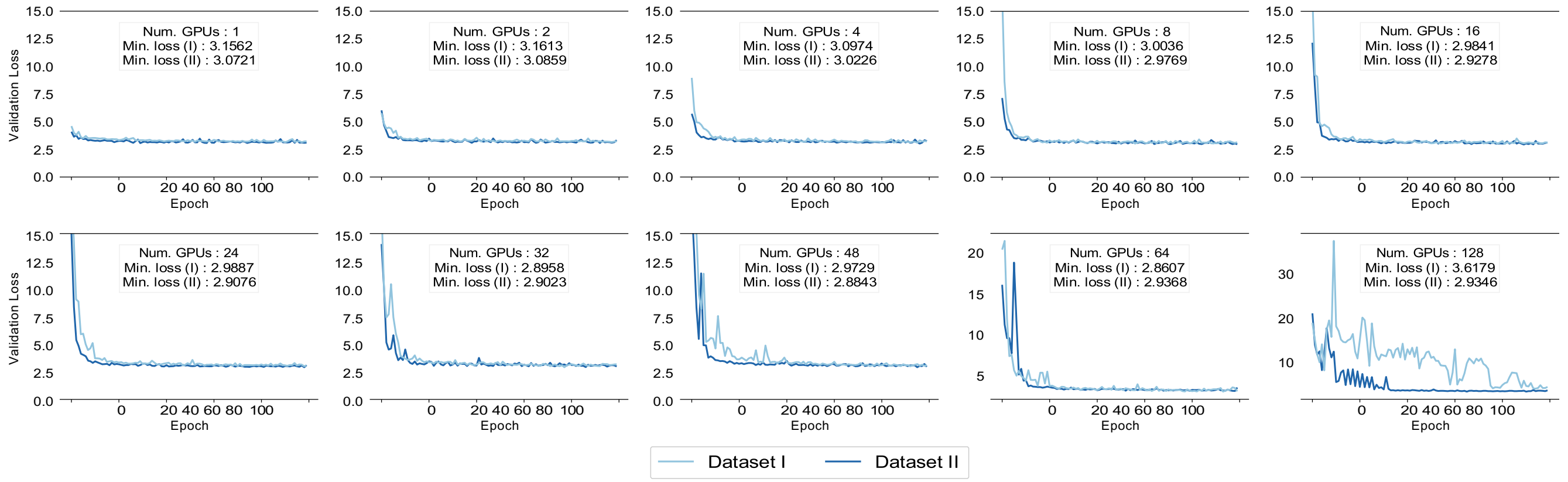

Fig. 5: Validation loss for the nowcast model trained for 100 epochs on two datasets as described in Section II-B. Divergence in the validation loss was avoided by adding warm-up epochs and lowering the learning to $\eta * \sqrt{N}$, where $\eta$ is the learning rate used on a single GPU and $N$ is the number of GPUs used. The minimum validation loss for Dataset I and II are shown on the respective figures. Y-axis limits for 64 and 128 GPUs is larger because of the significantly larger validation loss when training at these scales.

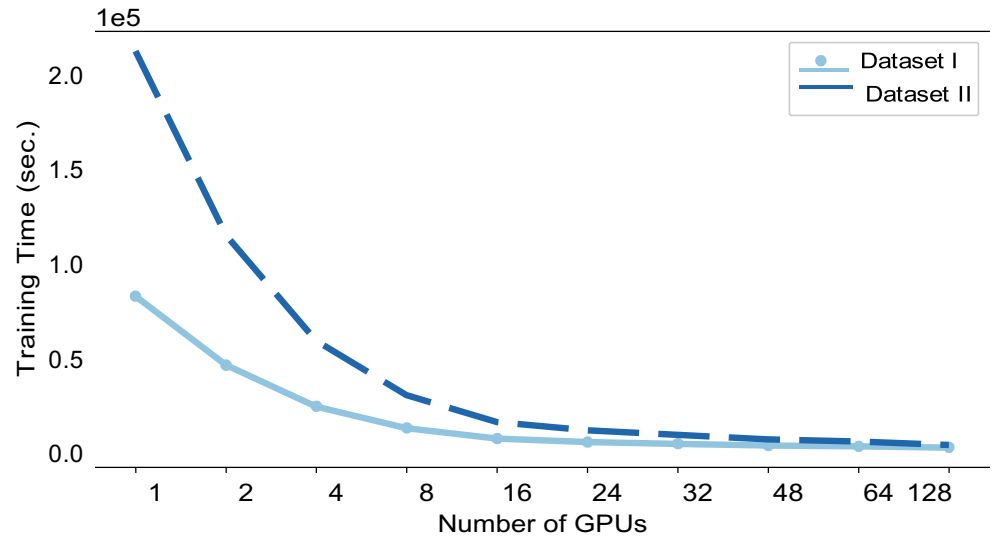

Fig. 6: Training time for the nowcast model on NVIDIA K80 GPUs across multiple compute nodes. The model was trained for 100 epochs.

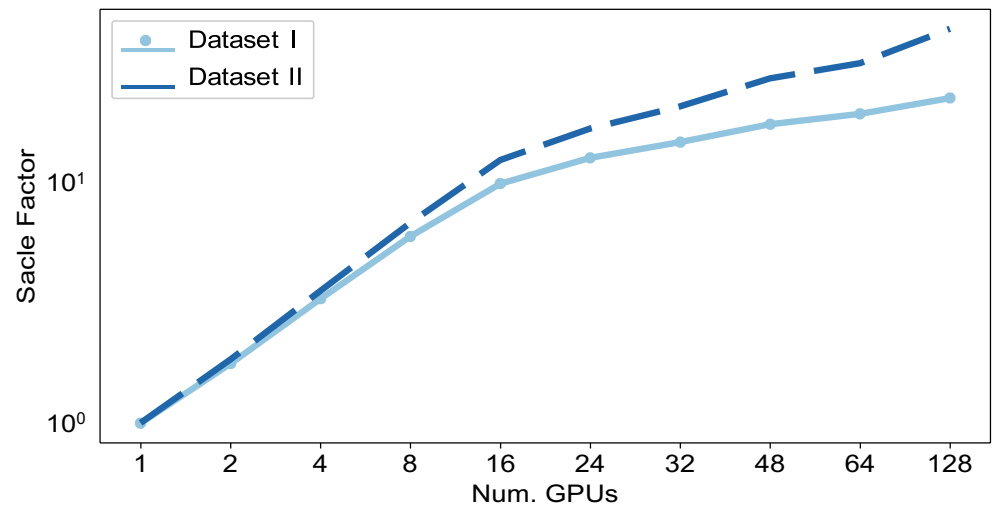

Fig. 7: Parallel speedup for the nowcast model trained for 100 epochs on NVIDIA K80 GPUs across multiple compute nodes.

Figure 7 shows the speedup observed as the number of GPUs used is increased. Here, it is observed that the speedup increases linearly as we add more GPUs, upto 16 GPUs. After this point, the communication costs start becoming more significant as compared with the compute time and the speedup observed is sub-linear. Due to the communication overhead, doubling the number of GPUs at each step, does not result in a doubling of the speedup as shown in Figure 8. For Dataset I, the maximum relative speedup of 1.862 is observed when the number of GPUs are doubled from 2 to 4. Correspondingly, for Dataset II, doubling GPUs from 2 to 4 and 4 to 8 provides a relative speedup of 1.928. After this point, each doubling of GPUs produces a speedup gain less than 1.8. This behavior is expected because the time cost of all-to-all communication for gradient averaging starts to dominate the computation time. If the model were more complex or the dataset larger, we can expect improved scaling with more GPUs. It is seen from Figure 7 that the training process scales better on Dataset II given the larger data size. This behavior is to be expected because of the data-parallel implementation of the training process.

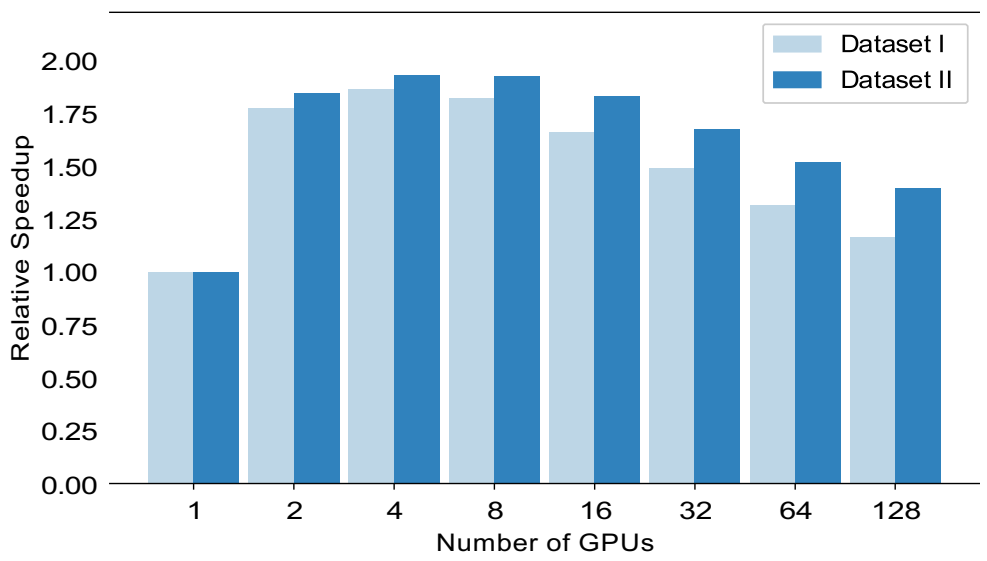

Fig. 8: Relative speedup of the Nowcast model training: Speedup for N GPUs is calculated using training time for previous GPU count. For example, relative speedup for 4 GPUs compares training time of 4 and 2 GPUs.

### C. *Initial performance of the nowcast model*

The MSE of the test set as a function of nowcast lead time is plotted in Figure 10. As a reference, the MSE of the persistence forecast (that is, the forecast generated by repeating the last frame of the input sequence for each future time) is also computed. This result shows the skill of the CNN significantly improves upon persistence, with the improvement becoming greatest at one hour.

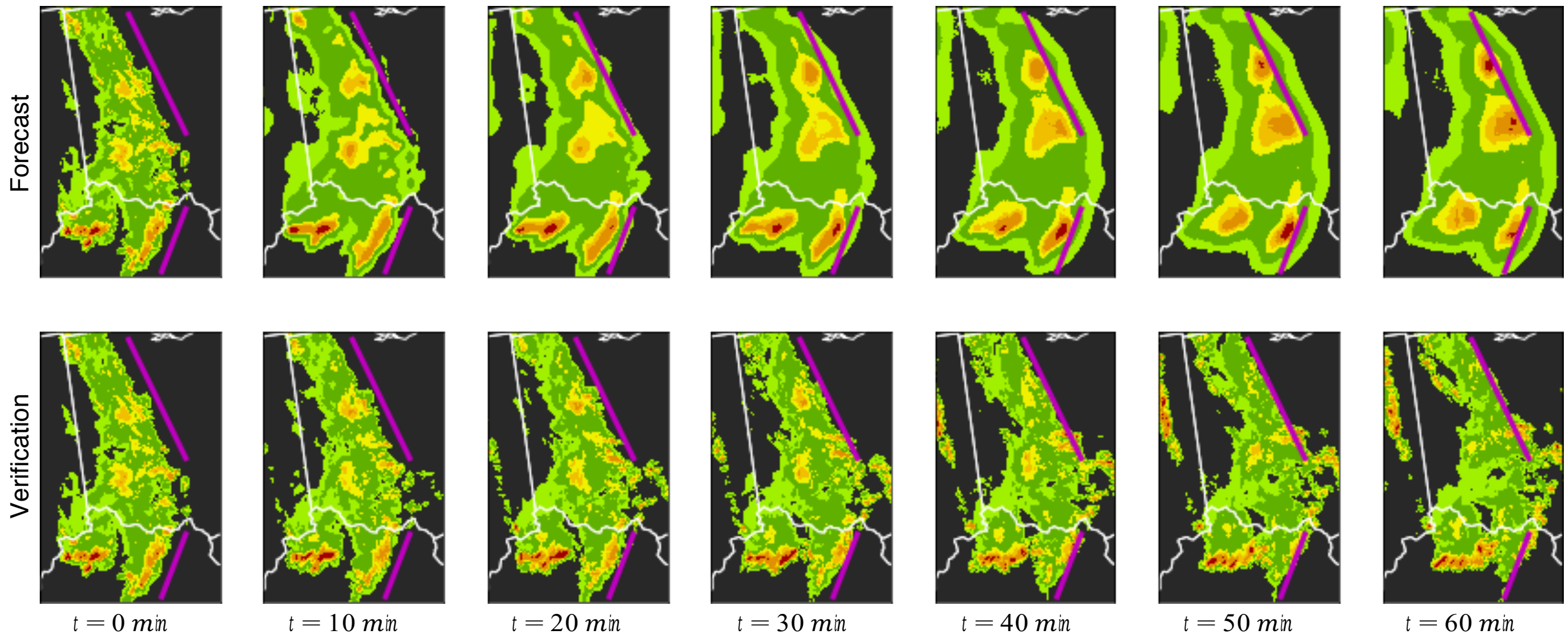

Fig. 9: A sample nowcast (top row) generated using the trained CNN alongside truth (bottom row). The frames represent 10 minute increments. The magenta lines are inserted as reference to see the movement of the storm. In this case, the eastward motion of the storm is captured by the CNN, with speed and position that is consistent with what is seen in the verification.

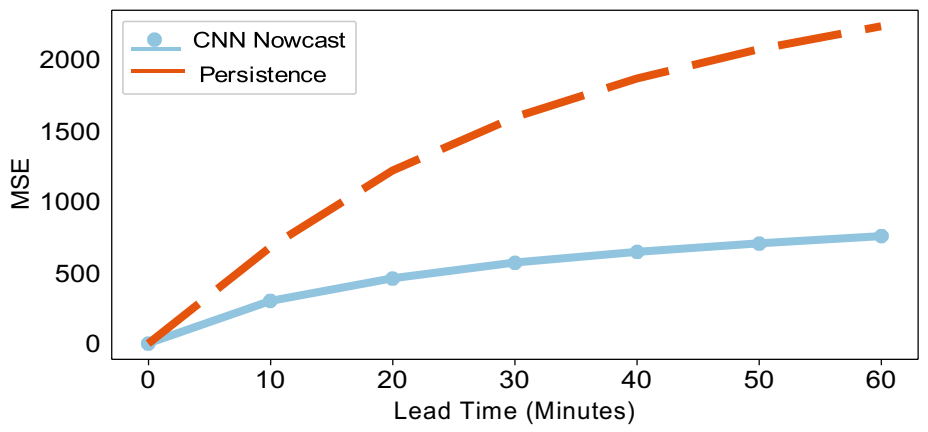

Fig. 10: Performance of the nowcast model as a function of lead time computed from a test set. The performance of the persistence forecast is also plotted as a reference.

The trained model was used to generate a nowcast of a convective line moving east across Illinois and Ohio on July 20th, 2018 at 1300 UTC. The results of the forecast are shown in Figure 9. In this case, the CNN is "moving" the storm system to the east, at a speed similar to that seen in the verification. The magenta lines are inserted east of the storm system as a reference.

As can be seen, the output of the CNN is relatively smoothed when compared to the verification. Since only MSE was used in the objective function, the high amount of spatial uncertainty of future weather leads to a solution that's somewhat blurred. While this leads to decreased MSE on average, from a nowcasting perspective this effect is not always ideal since the blur effect may appear to diminish smaller storms which may be confused with storm decay. To counteract this, a histogram matching procedure was applied to each forecast frame in Figure 9 to locally match the histograms of the forecast to the histogram of the initial condition. While this helped maintain some storm intensity throughout the 60 minute forecast, smaller scale features are still smoothed away in the longer leads. A better mitigation may be to expand the loss function in the neural network to also penalize this blurring and maintain perceptual similarity between the forecast and actual weather. This modification will continue to be explored in future work, thanks to the faster training enabled by distributed learning.

Another benefit of using CNNs for nowcasting is speed. The input image size for the test forecast in Figure 9 was 1000 x 1000 pixels, which was processed on a single workstation with 10 CPU cores in approximately 20 seconds. This is significantly faster than operational nowcasting systems, which require a cluster of CPUs and can take on the order of minutes to generate nowcasts. Thus if the performance of the CNN can be improved to match state of the art, using CNNs in operational nowcasting systems may be able to significantly reduce the time and resources required to generate these forecasts.

## V. Conclusion

In this paper we discuss the implementation of a data distributed, convolutional neural network for producing highly detailed short-term weather forecasts of precipitation. The nowcast model was developed in Tensorflow and parallelized across 128 NVIDIA K80 GPUs. Using the Horovod framework, we are able to reduce the training time of the model from over 2 days to just over an hour. For models of the size discussed in this paper, as few as 8 or 16 GPUs are sufficient to achieve significant speedups. The data distributed approach to parallel training offers the flexibility of scaling up from multiple GPUs on a single compute node to hundreds of distributed GPUs.

The use of distributed learning approaches enables weather researchers to develop, test and deploy new models significantly faster. This capability can enable better modeling of evolving weather patterns. Our current approach relies on small image tiles that are extracted from much larger images for training this model. As image tiles are increased in dimensions, we anticipate the model to perform better because of the availability of more local context for making predictions. However, because of the limited amount of GPU memory available, increasing tile sizes arbitrarily is not feasible. Future work includes evaluating this model on much larger image patches and a qualitative comparison between CPU and GPU implementations on this data.

## Acknowledgments

The authors acknowledge the MIT Lincoln Laboratory Supercomputing Center for providing HPC resources that have contributed to the research results reported in this paper. The authors thank Michael Jones on the LLSC team for his assistance with the machine learning software infrastructure setup.